%% file: preprint.tex
\documentclass[11pt,oneside]{article}

\usepackage[a4paper,
            bindingoffset=0.2in,
            left=1in,
            right=1in,
            top=1in,
            bottom=1in,
            footskip=.25in]{geometry}

\usepackage[english]{babel}
\usepackage[utf8]{inputenc}
\usepackage{amsmath,amsfonts}
\usepackage{graphicx}
\usepackage{xcolor}
\usepackage{soul}
\usepackage{multirow}
\usepackage{authblk}
\usepackage{booktabs}
\usepackage{url}

\title{Enhancing Brazilian Sign Language Recognition through Skeleton Image Representation}

\author{Carlos Eduardo G. R. Alves}
\author{Francisco de Assis Boldt}
\author{Thiago M. Paixão}
\affil[1]{Federal Institute of Espírito Santo, Campus Serra}
\affil[ ]{\normalsize cadu97@gmail.com, \{franciscoa,thiago.paixao\}@ifes.edu.br}
\date{}

\begin{document}
\maketitle

\begin{abstract}
Effective communication is paramount for the inclusion of deaf individuals in society. However, persistent communication barriers due to limited Sign Language (SL) knowledge hinder their full participation. In this context, Sign Language Recognition (SLR) systems have been developed to improve communication between signing and non-signing individuals. In particular, there is the problem of recognizing isolated signs (Isolated Sign Language Recognition, ISLR) of great relevance in the development of vision-based SL search engines, learning tools, and translation systems. This work proposes an ISLR approach where body, hands, and facial landmarks are extracted throughout time and encoded as 2-D images. These images are processed by a convolutional neural network, which maps the visual-temporal information into a sign label. Experimental results demonstrate that our method surpassed the state-of-the-art in terms of performance metrics on two widely recognized datasets in Brazilian Sign Language (LIBRAS), the primary focus of this study. In addition to being more accurate, our method is more time-efficient and easier to train due to its reliance on a simpler network architecture and solely RGB data as input.
\end{abstract}

\input{secs/1-intro}
\input{secs/2-method}
\input{secs/3-experiments}
\input{secs/4-results}
\input{secs/5-conclusion}

\bibliographystyle{ieeetr}
\bibliography{preprint}

\end{document}

%% file: secs/1-intro.tex
\section{Introduction}

Communication for deaf and hard-of-hearing individuals presents significant daily challenges due to the limited knowledge of Sign Language (SL) within the broader society. Globally, the deaf community comprises over 70 million, and this number is expected to increase dramatically. By 2050, it is estimated that around 700 million people worldwide will experience disabling hearing loss, representing approximately one in ten individuals worldwide \cite{Who2023, Wfdeaf}. These statistics underline the pressing need for greater support and enhanced accessibility to sign language, crucial steps towards fostering inclusive communication, and ensuring equal participation in society for all individuals, regardless of hearing ability.

SL is uniquely complex, relying on precise hand movements, facial expressions, and body language to convey meaning. Like speech, it has its own grammatical structures \cite{DeCastro2023}, accents, and dialects. Therefore, learning a SL entails mastering a second language, which is not always accessible to the general audience. To communicate with non-signers whether hearing or deaf, signing deaf individuals often resort to less natural and slower forms of communication, such as lip reading and/or written messages \cite{nunez2023survey}. Overcoming communication barriers necessitates, among other measures, technologies that facilitate SL learning and enable bidirectional translation: from signs to speech/text and vice versa.

To address these challenges, significant efforts have been dedicated to the development of Sign Language Recognition (SLR) systems. The objective is to recognize signs performed in front of a camera (and possibly other sensors) and translate them into textual form in a spoken language \cite{laines2023isolated}. According to Núñez-Marcos et al.  \cite{nunez2023survey}, there are two main variations of the SLR problem: Continuous Sign Language Recognition (CSLR) and Isolated Sign Language Recognition (ISLR). The key difference is that in ISLR, a video sequence comprises a single sign (word) rather than a sequence of connected signs. The focus of this study is on the recognition of isolated signs, which can facilitate sign translation given that signs are pre-segmented in an initial step. Nevertheless, the recognition of isolated signs also plays a crucial role in enabling gesture-based search engines \cite{Tamer2020KeywordSearch, Hassan2022AslSearch} or creating training platforms for hearing individuals to learn sign language at their own pace \cite{starner2024popsign}.

In the domain of ISLR, modern methodologies heavily rely on deep learning techniques for both feature extraction and sign classification. Various visual data modalities have been explored for SLR, including RGB frame content, depth maps, skeleton-based pose/facial information, thermal data, and motion flow \cite{Rastgoo2021}. Depth and pose information can be obtained directly from specialized sensors, such as Kinect V1 \cite{Cerna2021}, or, alternatively, extracted from RGB data. In the latter approach, there are works leveraging artificially generated depth maps \cite{DeCastro2023,sarhan2023pseudodepth} and pose/facial landmarks extracted from RGB frames by 3-rd party models \cite{DeCastro2023,lin2024skim}. For temporal information processing, popular choices include recurrent models based on Long Short-term Memory (LSTM) \cite{aly2020deeparslr,rastgoo2021hand} and Gated Recurrent Units (GRU) \cite{shen2024stepnet}, as well as 3-D Convolutional Neural Networks (3-D CNNs) \cite{li2020word,DeCastro2023,sarhan2020transfer}. Particularly, 3-D CNNs have gained popularity in addressing ISLR, largely due to the work of Sarhan and Frintrop \cite{sarhan2020transfer}. They proposed transferring spatiotemporal features from a 3-D CNN model (Inflated 3D, I3D) pre-trained on a large-scale action recognition dataset. 


Although 3-D CNNs have been becoming popular in ISLR, learning directly from RGB data may lead to complex 3-D CNN models with a large number of parameters \cite{laines2023isolated}. An alternative is to leverage skeleton-based representations due to their ability to better generalize to different scenarios \cite{laines2023isolated}, which has had a significant impact on the related task of Human Activity Recognition (HAR) \cite{memmesheimer2022skeleton}. In this direction, Yang et al. \cite{yang2018action} proposed converting temporal skeleton information into a 2-D image and processing it with a 2-D CNN to output the action label. Their method involves representing the skeleton as a tree and mapping the joint coordinates onto the image using a depth-first traversal of the tree structure. Recently, Laines et al. \cite{laines2023isolated} extended this idea to ISLR by incorporating facial key points and fine-grained hand joints, achieving state-of-the-art performance in two of the three tested datasets. Despite the promising results, as the authors themselves claim, the use of skeleton images and 2-D CNNs is relatively underexplored in the ISLR literature.

Motivated by the aforementioned observations, we propose to advance in this direction by investigating the utilization of skeleton image representation and 2-D CNNs for recognizing isolated signs in the Brazilian Sign Language (LIBRAS). Our approach leverages robust extraction of body, hands, and face key points (landmarks) delivered by OpenPose \cite{cao2019openpose} and the Skeleton-DML method \cite{memmesheimer2022skeleton} for image representation due to its high performance on HAR. Notably, our method surpasses the state-of-the-art (multimodal 3-D CNN \cite{DeCastro2023}) on both MINDS-Libras \cite{Rezende2021} and LIBRAS-UFOP \cite{Cerna2021}, which are the most widely recognized LIBRAS datasets for ISLR. Additionally, the proposed method is more time-efficient and easier to train.

In summary, the main contributions of this paper are:
\begin{itemize}
    \item The introduction of a simple yet efficient method for recognizing isolated signs using skeleton image representation.
    \item An ablation study elucidating the essential components of the proposed method.
    \item A comparative study, where our method achieves state-of-the-art performance on the two most popular LIBRAS datasets: MINDS-Libras \cite{Rezende2021} (accuracy of 0.93) and LIBRAS-UFOP \cite{Cerna2021} (accuracy of 0.82).
\end{itemize}

The rest of this paper is structured as follows. Section \ref{sec:method} introduces the proposed method. Section \ref{sec:experiments} describes the experimental methodology, and the results are presented and discussed in Section \ref{sec:results}. Finally, the conclusion and future work are discussed in Section \ref{sec:conclusion}. 

%% file: secs/2-method.tex
\begin{figure}[t]
  \centering
  \includegraphics[width=0.8\linewidth]{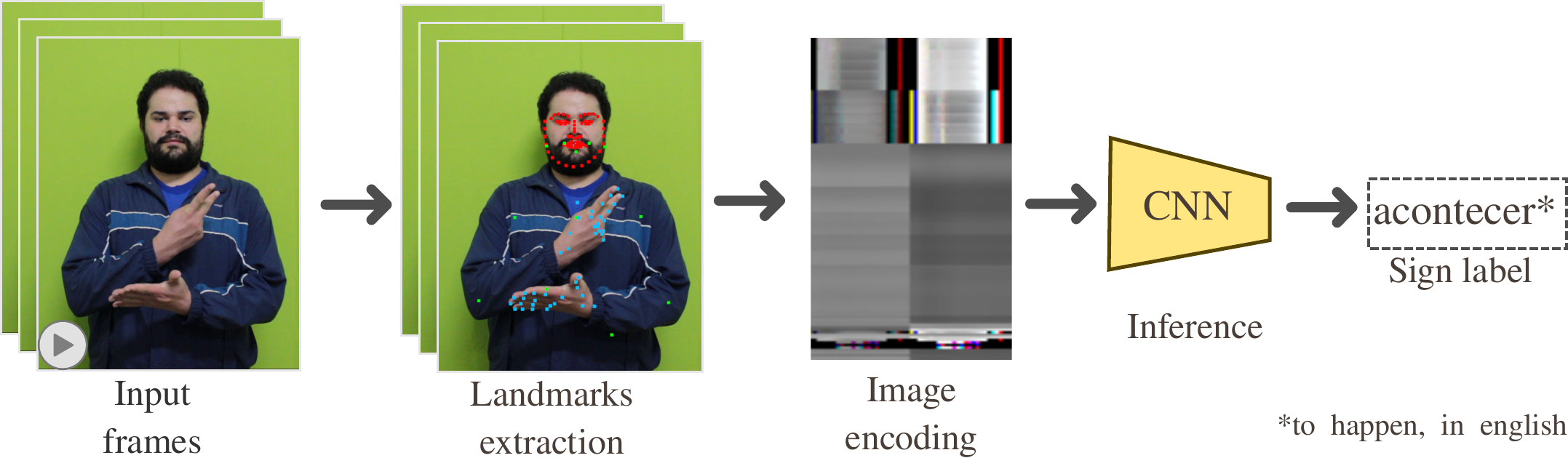}
  \caption{Overview of the proposed ISLR method. Initially, landmarks are extracted from individual frames of the input video sequence. Then, the landmarks are converted into a single 2-D image that encodes spatial and temporal information. Finally, the image is fed into a CNN model for sign classification.}
  \label{fig:overview}
\end{figure}

\section{Proposed ISLR Method}
\label{sec:method}

The basic pipeline of the proposed method is illustrated in Figure \ref{fig:overview}. Initially, the frames of the input video sequence, comprising a single sign, are processed individually to extract body, hands, and face landmarks. Each landmark point consists of three components: the 2-D spatial coordinates and the time position, corresponding to the frame from which it was extracted. Subsequently, the landmarks are converted into a single 2-D image that encodes both spatial and temporal information. This image is then fed into a CNN model, which outputs the class label of the sign with maximum probability (sign label). The following sections describe each part of the presented pipeline, as well as the process to train the CNN model.

\subsection{Landmarks Extraction}
\label{sec:landmarks-extraction}

\begin{figure}[t]
  \centering
  \includegraphics[width=0.8\linewidth]{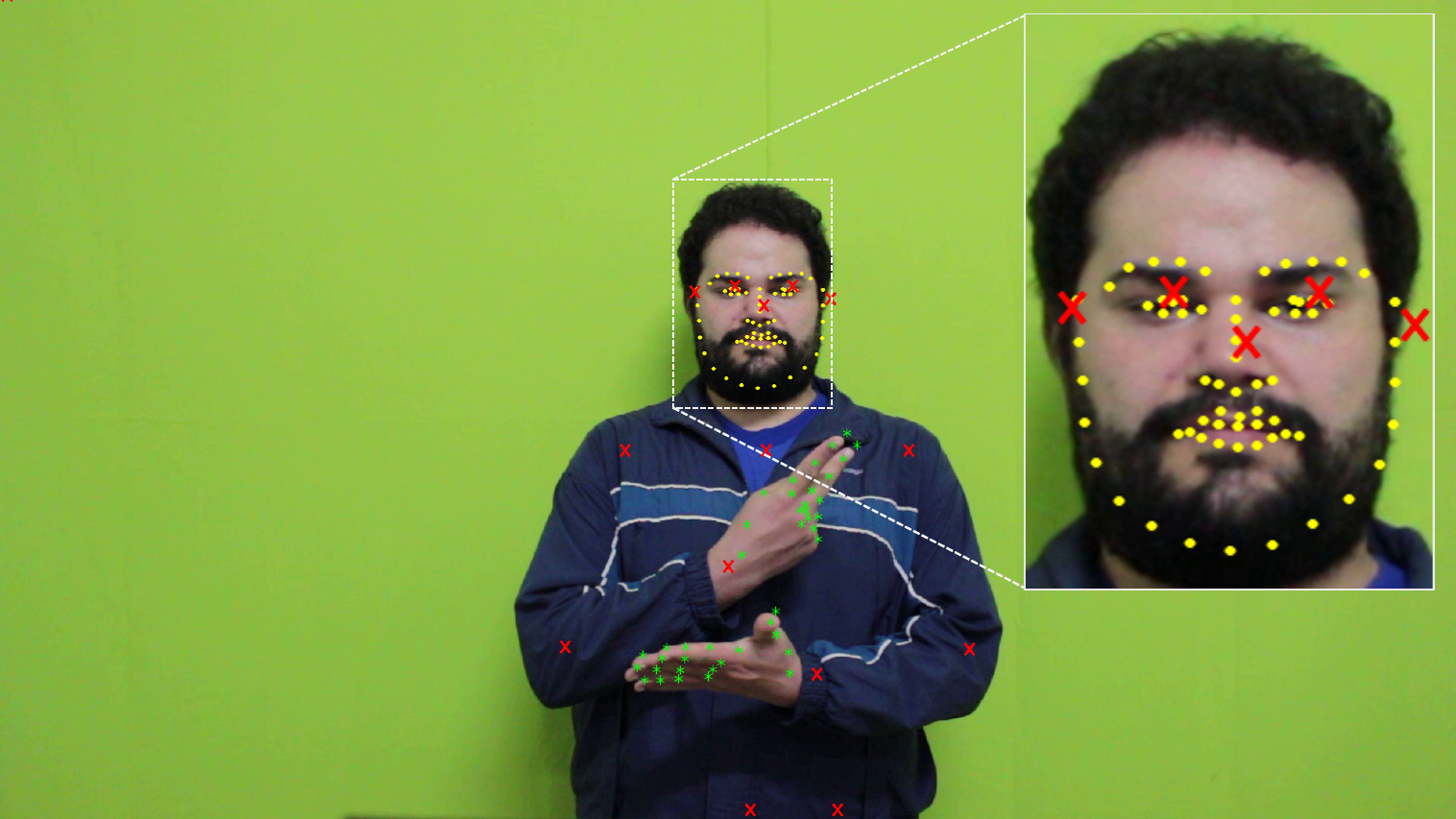}
  \caption{Extraction of landmarks from raw RGB frames using OpenPose \cite{cao2019openpose}. All landmarks above the hip (totaling 126) were utilized, including 42 hand landmarks (green dots), 70 face landmarks (yellow dots), and 14 body landmarks (red `x'). Face points indicated by `x' markers are treated as body landmarks by OpenPose.}
  \label{fig:landmarks}
\end{figure}

In this initial stage, the space-time landmarks are extracted from the raw RGB frames of the input video sequence. This extraction process relies on OpenPose \cite{cao2019openpose}, a deep learning-based open-source library for estimating human poses by extracting 2-D keypoints. OpenPose has been widely used in various applications, including movement analysis in sports \cite{chen2023research} and physiotherapy \cite{saiki2023reliability}, as well as human activity recognition \cite{emanuel2021snapshot}. Due to its capability to detect facial features, which are crucial for Sign Language Recognition (SLR), OpenPose is also utilized to incorporate facial elements into machine learning models for SLR \cite{DeCastro2023}.

To capture movement and facial features, all landmarks above the hip were utilized, which corresponds to 126 landmarks from a total of 137 provided by OpenPose. As depicted in Figure \ref{fig:landmarks}, there are 42 points representing the landmarks of both hands, denoted by the green dots, 70 yellow points correspond to face landmarks, and 14 red `x' markers indicating body landmarks (the 5 points situated in the face are treated as body landmarks by OpenPose). Finally, the depth coordinate inferred by OpenPose was discarded since our method focuses on 2-D data.

\subsection{Image Encoding}
\label{sec:image_transformation}

The image encoding is achieved through the Skeleton-DML algorithm \cite{memmesheimer2022skeleton}. Although originally intended for representing skeleton joints for human activity recognition, the underlying concept applies to space-time coordinates in general, as is the case with the landmarks in our method. Additionally, adapting to 2-D coordinates is straightforward, as omitting the depth coordinate (z-axis) simply implies waiving a portion of the image data. The following paragraph describes more formally the encoding process.

Given a video sequence with $T$ frames, the set of extracted landmarks for a frame $j \in \{1, 2, \ldots, T\}$ can be represented as coordinates $\{(x^{(j)}_i,~y^{(j)}_i)\}_{i=1}^{L}$, where $L=126$ is the total number of landmarks. The $x$ and $y$ coordinates are real values normalized into the $[0,1]$ range. In the encoding process, the coordinates are converted into the matrices $\mathbf{X}=[X_{ij}]_{L \times T}$ and $\mathbf{Y}=[Y_{ij}]_{L \times T}$, where $X_{ij} = x^{(j)}_i$ and $Y_{ij} = y^{(t)}_i$. The matrices are reshaped into $L \times T / 3 \times 3$ tensors $\mathbf{X}'$ and $\mathbf{Y}'$, so that $X'_{ijk} = X_{i,3j+k}$ and $Y'_{ijk} = Y_{i,\,3j+k}$.
Finally, the encoded image corresponds to $I = \mathbf{X}' \oplus \mathbf{Y}'$, where $\oplus$ denotes the concatenation along the horizontal axis. The values of $I$ are scaled to integer values in the $[0, 255]$ range. Notice that $I$ is a 3-channel (RGB) image with $L$ rows and $2T/3$ columns.

It is worth mentioning that the interpretation of the generated image is not straightforward. As shown in Figure \ref{fig:overview}, dark and bright pixels indicate the leftmost and rightmost positions in the frame, respectively. Furthermore, constant rows indicate that the position of the associated landmark is not changing considerably. Colored pixels, on the other hand, represent fast movements, indicating sharp variations in the horizontal and/or vertical direction within a 3-length temporal window.

\subsection{Training the Model}
\label{sec:model}

The training of the model leverages a collection of annotated videos, each one comprising a single sign. In a preliminary (offline) step, each video sequence is processed to extract and store the respective landmarks. In each training epoch, the landmarks undergo an augmentation procedure before being converted into 2-D images to train the convolutional model. The augmentation process aims to enhance data diversity across training epochs by applying basic transformations, including rotation, zoom, translation, and horizontal flip. In our approach, this process does not increase the number of samples.

The adopted CNN architecture consists of an 18-layer Deep Residual Network (ResNet18) \cite{he2016deep} (base network), followed by two fully connected layers: a 128-unit layer with ReLU activation for feature extraction, and an output layer whose size matches the number of signs (classes) in the dataset. The input size of the network is $224 \times 224$, while the encoded images have dimensions $126 \times 2T/3$. Although the length of the video sequences $T$ may vary across the samples, the encoded image width ($2T/3$) is consistently lower than 224 for the evaluated cases. Indeed, having $L \geq 224$ would imply shot lengths of approximately 11 seconds (assuming a frame rate of 30 fps), which is unlikely for isolated signs. Therefore, there is no loss of information when resizing the image to fit the network input size because the image dimensions are always increased.

During training, the base network is initialized with pre-trained weights from ImageNet. The model undergoes training for 20 epochs using the Adam algorithm to minimize cross-entropy loss, with a mini-batch size of 64, and a learning rate of 0.0001. An early stopping mechanism is implemented, halting training if the validation loss fails to decrease over five consecutive epochs. To mitigate overfitting, batch normalization is applied to the output of the base network, and dropout regularization with a probability of 0.5 is employed after the 128-unit layer. Furthermore, L2-regularization with a weight of 0.0001 is applied to the network parameters. The model state of the epoch that achieved the highest validation accuracy was chosen as the ``best model'' for posterior evaluation.

%% file: secs/3-experiments.tex
\begin{figure}[t]
  \centering
  \includegraphics[width=\linewidth]{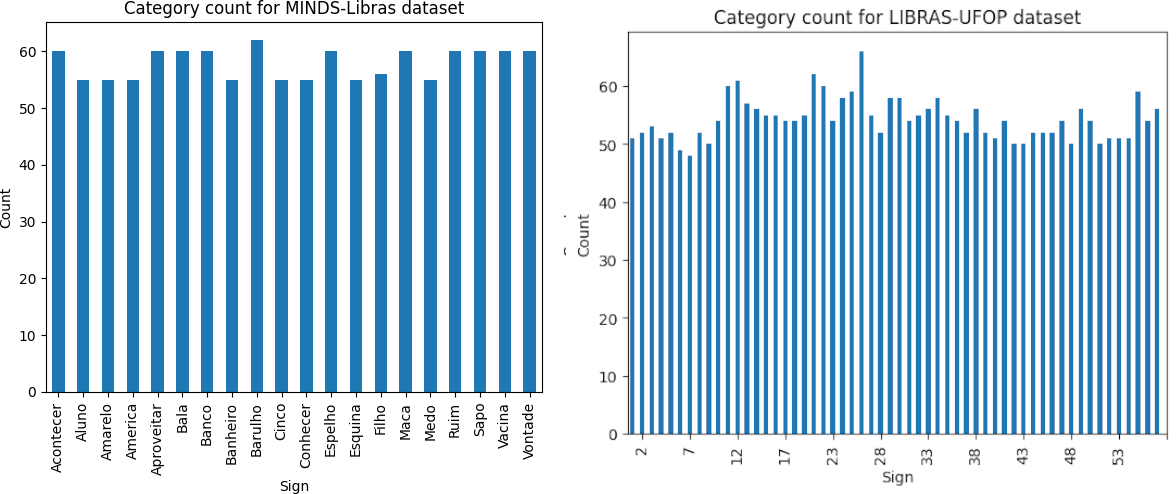}
  \caption{Distribution of video sequence length per sign for the evaluation datasets.}
  \label{fig:sig_count_hist}
\end{figure}

\section{Experimental Methodology}
\label{sec:experiments}

This section provides a comprehensive overview of the methodology employed in this investigation, covering the selection and preparation of public datasets, the choice of performance metrics for model evaluation, and the details of the experimental procedure.

\subsection{Datasets}
\label{sec:datasets}

The datasets utilized in our experiments comprise video sequences depicting isolated signs in Brazilian Sign Language (LIBRAS). We leveraged two prominent public collections for this purpose: the MINDS-Libras dataset \cite{Rezende2021} and the LIBRAS-UFOP dataset \cite{Cerna2021}. Our literature review indicates that these datasets play a central role in ISLR in LIBRAS, enabling us to compare our method to the literature. Figure \ref{fig:sig_count_hist} shows the distribution of video sequence length per sign for both datasets. The charts demonstrate that both collections are reasonably balanced, ensuring that the number of data points for each class is approximately equal.

\paragraph{MINDS-Libras Dataset} This collection includes a subset of 20 distinct signs in LIBRAS selected by a specialist. The signs were selected by their phonological characteristics, which include aspects like hand configuration, location and movements, palm orientation, and facial expressions. They were performed by 12 signers, each signer executed the same sign five times resulting in a total of 1,158 video sequences. This number falls slightly short of the expected 1,200 samples due to the absence of certain signs from four signers. The performed signs (classes) include: ``To happen'', ``Student'', ``Yellow'', ``America'', ``To enjoy'', ``Candy'', ``Bank'', ``Bathroom'', ``Noise'', ``Five'', ``To know'', ``Mirror'', ``Corner'', ``Son'', ``Apple'', ``Fear'', ``Bad'', ``Frog'', "Vaccine'', and ``Will''.

\paragraph{LIBRAS-UFOP Dataset} This collection encompasses a subset of 56 distinct signs in LIBRAS performed by 5 signers. As detailed in \cite{DeCastro2023}, the signs are classified into four categories based on the similarity level of movement, articulation point, hand pose, and facial expression (a sign may belong to more than one category): (i) same movement and articulation point but different hand pose; (ii) different movements but the same articulation point and hand configuration (19 signs); (iii) same movement and hand configuration but different articulation points (8 signs); (iv) similar movement but different facial expression (19 signs). The similarity between signs presents a challenging factor in this dataset.
In contrast to MINDS-Libras, each video recording contains from 8 to 16 takes of the same sign performed sequentially. To accurately segment the time frame of each sign sample, we used an auxiliary labeling file provided along the dataset. This file specifies the precise frame where each sign begins and ends for every repetition across all videos, ensuring precise delineation of sign boundaries. Unlike MINDS-Libras, the time window of a sign in LIBRAS-UFOP does not encompass frames in which the signer is at rest position. To ensure consistency between both datasets, we included the 15 frames preceding and following the annotated frames.

\subsection{Performance Metrics}

The performance analysis leveraged four commonly used metrics in the SLR literature: accuracy, precision, recall, and F1-score. Accuracy measures the overall correctness of the model as the ratio of correct predictions to total predictions. Precision assesses the accuracy of positive predictions, calculated as the ratio of true positives to all positive predictions. Recall, or sensitivity, quantifies the model's ability to identify all relevant instances, calculated as the ratio of true positives to actual positives. The classes are weighted equally in the metrics computation, therefore, we reported the macro average of precision and recall. F1-score harmonizes precision and recall into a single metric, useful for comparing models with similar accuracies but different precision and recall values.

\subsection{Experimental Procedure}
\label{sec:exp-procedure}

The basic experimental procedure to evaluate the efficacy of the proposed method is described as follows. In a preliminary stage, the body, hand, and face landmarks were extracted from the raw video sequences, as described in Section \ref{sec:landmarks-extraction}. We assume that, at this point, the signs in the LIBRAS-UFOP dataset were already individualized according to the procedure described in Section \ref{sec:datasets}. The following steps are applied for each dataset individually.

The partitioning of train-test samples adheres to a nested \emph{Leave One Person Out} (LOPO) cross-validation protocol, reflecting real-world application scenarios where the recognition system must operate without prior knowledge of the current signer. Following this approach, the train partition was further subdivided into exclusive train and validation sets, with each of the remaining signers used once to compose the validation set. This set was used to determine the best model, as discussed in Section \ref{sec:model}. Consequently, for a dataset comprising $n$ signers, a total of $n(n-1)$ train-test sections were conducted, ensuring thorough evaluation and robustness of the recognition system across diverse signer profiles.

In each training epoch of an experiment (section), the landmarks designated exclusively for training underwent the augmentation procedure before being converted into images, as detailed in Section \ref{sec:model}. Conversely, validation and test landmarks in each experiment were only converted into images once. Following the training phase, the model was evaluated on the corresponding test set, and the performance metrics were recorded.



\paragraph{Ablation Study} This study aims to assess the impact of excluding data augmentation from the training stage process, along with the potential benefits of implementing a mechanism to uniformize the number of frames per video sequence. The uniformization mechanism, adapted from \cite{da2024multiple}, involves setting a target number of frames per dataset and adjusting the frame count of each video sequence accordingly. Specifically, if a video contains fewer frames than the target, its last frame is duplicated until the desired count is reached. Conversely, if a video exceeds the target, frames are removed at spaced intervals to achieve the desired total, ensuring consistency across the dataset. The target frame count is determined as the average number of frames across the dataset. This study follows the outlined experimental procedure, with the difference that each investigated component is individually enabled or disabled (one-at-a-time approach). The resulting outcomes are then compared to the proposed method configuration, where augmentation is enabled and no uniformization mechanism is implemented.

\paragraph{Comparative Evaluation} The proposed method was compared to the state-of-the-art multi-stream 3D CNN architecture developed by De Castro et al. \cite{DeCastro2023}. This model integrates segmented hands and faces, pose information, speed maps, RGB images, and artificially generated depth maps. It achieved superior accuracy compared to the machine learning models developed by Passos et al. \cite{Passos2021} on MINDS-Libras and LIBRAS-UFOPS, and outperformed the multimodal approach of Cerna et al. \cite{Cerna2021} on LIBRAS-UFOP.



\subsection{Experimental Platform}

The experimental platform utilized Ubuntu 18.04.5 operating through Windows Subsystem for Linux (WSL) on a Windows 11 platform. The hardware configuration included an Intel i7 13700K processor clocked at 5.4GHz, 32GB of DDR4 RAM operating at 3200MT/s, and an NVidia RTX 4070 GPU. The development of the proposed neural network was carried out using Python v3.9.16, PyTorch 2.0.1, and CuDNN version 11.8.
The source code and pre-trained models will be publicly available upon acceptance of this work.

%% file: secs/4-results.tex
\begin{table}[t]
    \centering
   
    \resizebox{0.8\linewidth}{!}{%
    \begin{tabular}{ccccc}
        \toprule
        Dataset      & Accuracy        & Precision       & Recall          & F1-score        \\
        \midrule
        MINDS-Libras & 0.93 $\pm$ 0.05 & 0.94 $\pm$ 0.05 & 0.93 $\pm$ 0.05 & 0.93 $\pm$ 0.05 \\
        LIBRAS-UFOP  & 0.82 $\pm$ 0.04 & 0.83 $\pm$ 0.05 & 0.81 $\pm$ 0.05 & 0.80 $\pm$ 0.05 \\
        \bottomrule
    \end{tabular}
    }
     \caption{Results of the proposed method in its default configuration (avg. metric $\pm$ standard deviation).}
    \label{tab:default}
\end{table}

\section{Results and Discussion}
\label{sec:results}

Table \ref{tab:default} shows the results obtained with the proposed method in the default configuration. For the MINDS-Libras dataset, the proposed method achieved an accuracy of 0.93, precision of 0.94, recall of 0.93, and F1-score of 0.93. For LIBRAS-UFOP, accuracy was 0.82, with a precision of 0.83, recall of 0.81, and F1-score of 0.82. The remarkable difference in performance is consistent with findings in the literature \cite{DeCastro2023,Passos2021}, highlighting LIBRAS-UFOP as a more challenging dataset.

\begin{figure}[t]
  \centering
  \includegraphics[width=0.7\linewidth]{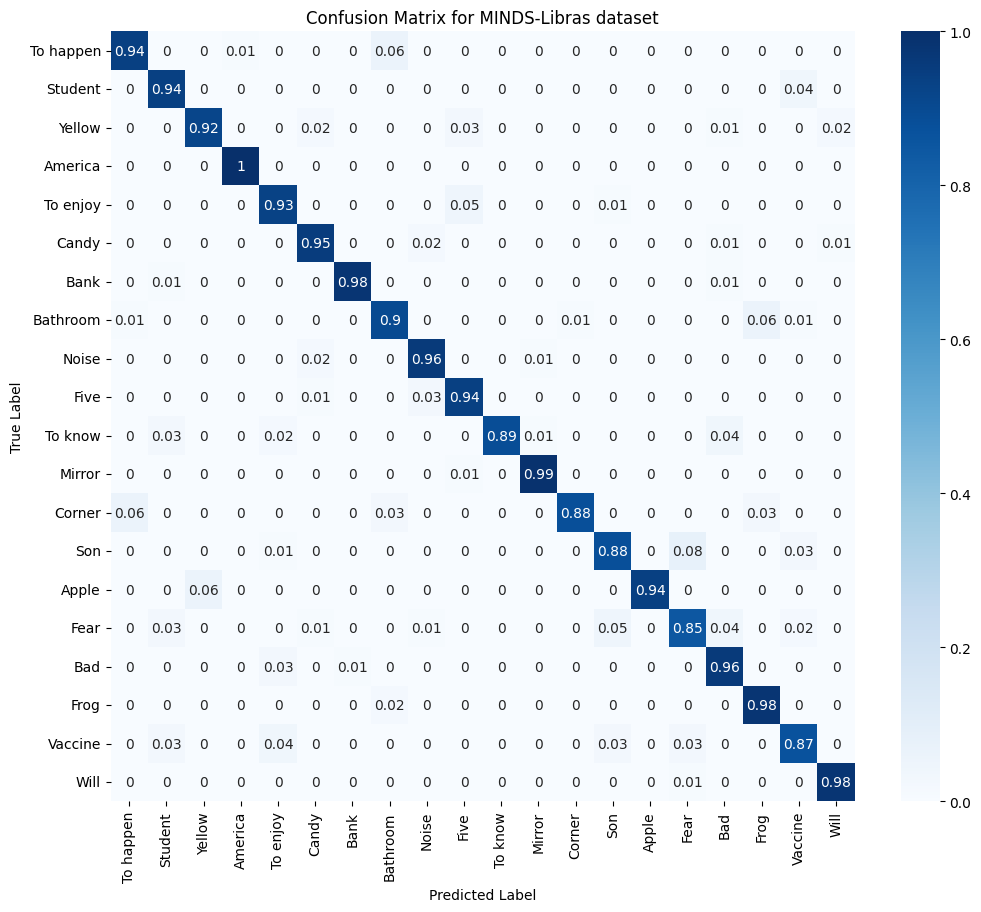}
  \caption{Confusion matrix for the MINDS-Libras dataset.}
  \label{fig:confusion_matrix_minds}
\end{figure}

The confusion matrix in Figure \ref{fig:confusion_matrix_minds}, normalized by true values (row values), offers an in-depth analysis of the recognition performance within the MINDS-Libras Dataset. The y-axis denotes the true label for each sign, while the x-axis reflects the labels predicted by our model. The main diagonal represents the precision for each sign. Out of 20 signs, 15 achieved a precision of at least 0.9, whereas only one sign, ``Fear'', registered a precision of 0.85. This sign was frequently misidentified as ``Son'',  ``Bad'', or ``Student''. The best performance was observed for ``America'', with a precision of 1.

\begin{table}[t]
    \centering

    \resizebox{\linewidth}{!}{%
    \begin{tabular}{ccccccc}
        \bottomrule
        Dataset & Data aug. & Uniform. & Accuracy & Precision  & Recall & F1-score \\
        \midrule
        \multirow{3}{*}{MINDS-Libras} & \checkmark & & 0.93 $\pm$ 0.05 & 0.94 $\pm$ 0.05 & 0.93 $\pm$ 0.05 & 0.93 $\pm$ 0.05 \\
                                      & \checkmark & \checkmark & 0.88 $\pm$ 0.08 & 0.89 $\pm$ 0.06 & 0.87 $\pm$ 0.08 & 0.86 $\pm$ 0.09 \\
                                      & & & 0.92 $\pm$ 0.06 & 0.93 $\pm$ 0.06 & 0.92 $\pm$ 0.06 & 0.91 $\pm$ 0.07 \\
        \midrule
        \multirow{3}{*}{LIBRAS-UFOP} & \checkmark & & 0.82 $\pm$ 0.04 & 0.83 $\pm$ 0.05 & 0.81 $\pm$ 0.05 & 0.80 $\pm$ 0.05 \\
                                     & \checkmark & \checkmark & 0.85 $\pm$ 0.03 & 0.85 $\pm$ 0.04 & 0.84 $\pm$ 0.04 & 0.82 $\pm$ 0.04 \\
                                     & & &  0.82 $\pm$ 0.04 & 0.82 $\pm$ 0.05 & 0.81 $\pm$ 0.04 & 0.79 $\pm$ 0.05 \\
        \bottomrule
    \end{tabular}}
        \caption{Results of the ablation study (avg. metric $\pm$ standard deviation). The second and third columns specify the enabled features. The first row for each dataset represents the method in its default configuration.}
        \label{tab:ablation_results}
\end{table}

\subsection{Ablation Study}
\label{sec:ablation_study}

Table \ref{tab:ablation_results} presents the results of the ablation study conducted as outlined in Section \ref{sec:exp-procedure}. The second and third columns specify the features present in the training, with the first row for each dataset representing the method in its default configuration.
As observed in the table, the data augmentation had a modest positive impact. The most significant influence on performance was observed for the frames uniformization mechanism, which increased accuracy for LIBRAS-UFOP by 3 percentage points (p.p.). However, it had the opposite effect on MINDS-Libras, resulting in a decrease in accuracy by 5 p.p.

\subsection{Comparative Evaluation}

Table \ref{tab:results} provides the results for the comparison with the state-of-the-art on the evaluated datasets. The results of the compared methods correspond to those reported in \cite{DeCastro2023}, and the ``-'' symbol indicates entries for which the metric was not available.
Overall, the proposed method achieved the best performance in all the available metrics. In MINDS-Libras, it outperformed \cite{DeCastro2023} (state-of-the-art) by 2 p.p. in accuracy and 3 p.p. in F1-score, with a lower standard deviation in both metrics. The remarkable point is that our method stands out for its simplicity compared to the complex multimodal 3-D CNN approach proposed in \cite{DeCastro2023}. For the LIBRAS-UFOP dataset, our method outperformed \cite{DeCastro2023} and \cite{Cerna2021} by a larger margin: 8 p.p. in accuracy and 9 p.p. in F1-score.

\begin{table}[t]
    \centering
    \resizebox{\linewidth}{!}{%
    \begin{tabular}{cccccc}
        \toprule
        Method & Dataset & Accuracy & Precision & Recall & F1-score \\
        \midrule
        \textbf{Ours} & \multirow{3}{*}{MINDS-Libras} & \textbf{0.93  $\pm$ 0.05} & \textbf{0.94 $\pm$ 0.05} & \textbf{0.93 $\pm$ 0.05} & \textbf{0.93 $\pm$ 0.05} \\
        De Castro (2023) \cite{DeCastro2023}  &  & 0.91 $\pm$ 0.07 & - & -          & 0.90 $\pm$ 0.08 \\
        Passos et al. (2021) \cite{Passos2021} &  & 0.85 $\pm$ 0.02 & -             & -          & -             \\
        \midrule
        \textbf{Ours} & \multirow{4}{*}{LIBRAS-UFOP} & \textbf{0.82 $\pm$ 0.04} & \textbf{0.83 $\pm$ 0.05} & \textbf{0.81$\pm$ 0.05} & \textbf{0.80 $\pm$ 0.05} \\
        De Castro (2023) \cite{DeCastro2023} & & 0.74 $\pm$ 0.04 & -             & -          & 0.71 $\pm$ 0.05 \\
        Passos et al. (2021) \cite{Passos2021} & & 0.65 $\pm$ 0.04 & -             & -          & -             \\
        Cerna et al. (2021) \cite{Cerna2021} & & 0.74 $\pm$ 0.03 & -             & -          & -             \\ 
        \bottomrule
    \end{tabular}}
    \caption{Results of the comparative evaluation (avg. metric $\pm$ standard deviation). Values highlighted in \textbf{bold} indicate the best performance for a given dataset. The ``-'' marker indicates that the metric is not available.}
\label{tab:results}
\end{table}

\paragraph{Limitations on Time Efficiency}

A limitation of our approach is the time burden of extracting the landmarks, which in turn is due to the dependency on the OpenPose library. On average, processing a video sequence with OpenPose takes nearly 36 seconds, resulting in an approximate frame rate of 2 frames per second, hindering real-time performance. Nonetheless, the same issue is observed in the method of De Castro et al. \cite{DeCastro2023}. Their 3-D CNN achieves its highest accuracy by utilizing hand and face pose data extracted by OpenPose. Without this data, the accuracy decreases dramatically by more than 25 p.p., as reported by the authors. The rest of our pipeline takes, on average, around 4.58 milliseconds to classify the entire sequence.

%% file: secs/5-conclusion.tex
\section{Conclusion}
\label{sec:conclusion}

In this paper, we proposed a method to recognize isolated signs in SL based on skeleton image representation and classification via 2-D CNN. Compared to the state-of-the-art, typically based on multimodal 3-D CNNs, our approach is more time-efficient and easier to train since it relies on a simpler network architecture and solely on RGB data as input. This approach facilitates integration into everyday technologies, enhancing communication for the deaf and hard-of-hearing individuals in various settings.

The ablation study evaluated the performance of our method after disabling data augmentation and including a mechanism to uniformize the video sequence length. Results showed the impact of the adopted augmentation strategy was negligible. However, non-consistent results were observed for the uniformization of the video sequence length: there was a decrease in accuracy of 5 p.p. for MINDS-Libras and an increase of 3 p.p. for LIBRAS-UFOP. This fact deserves further investigation, which is beyond the scope of this paper. Additionally, our method, in its default configuration, outperformed competitive methods on the two evaluated datasets in all reported metrics, achieving accuracies of 0.93 and 0.82 for MINDS-Libras and LIBRAS-UFOP, respectively. Particularly for LIBRAS-UFOP, this represents a remarkable increase of 8 p.p. in accuracy compared to state-of-the-art \cite{DeCastro2023}.

Nonetheless, the time burden imposed by OpenPose in landmarks extraction is still a limiting factor, as it also happens to \cite{DeCastro2023}. To tackle this issue, alternative pose extraction tools, such as MMPose \cite{mmpose2020} and MediaPipe \cite{lugaresi2019mediapipe}, will be carefully investigated. Beyond simply replacing tools, we intend to develop strategies to address the typically less accurate detection observed in faster tools. Furthermore, future work will explore other image encoding algorithms and CNN models, and also investigate transfer learning across different sign languages or even across different (but correlated) tasks, such as human activity recognition.